# A novel transformer-based approach for soil temperature prediction

Muhammet Mücahit Enes **Yurtsever**[a], Ayhan **Küçükmanisa**[b] and Zeynep Hilal **Kilimci**[a,*]

[a]*Department of Information Systems Engineering, Kocaeli University, 41001, Kocaeli, Turkey*
[b]*Department of Electronics and Communication Engineering, Kocaeli University, 41001, Kocaeli, Turkey*

## ABSTRACT

Soil temperature is one of the most significant parameters that plays a crucial role in glacier energy, dynamics of mass balance, processes of surface hydrological, coaction of glacier-atmosphere, nutrient cycling, ecological stability, the management of soil, water, and field crop. In this work, we introduce a novel approach using transformer models for the purpose of forecasting soil temperature prediction. To the best of our knowledge, the usage of transformer models in this work is the very first attempt to predict soil temperature. Experiments are carried out using six different FLUXNET stations by modeling them with five different transformer models, namely, Vanilla Transformer, Informer, Autoformer, Reformer, and ETSformer. To demonstrate the effectiveness of the proposed model, experiment results are compared with both deep learning approaches and literature studies. Experiment results show that the utilization of transformer models ensures a significant contribution to the literature, thence determining the new state-of-the-art.

## 1. Introduction

The significance of soil temperature within the context of environmental dynamics is underpinned by its intricate influence on a multitude of ecological processes, rendering it a fundamental determinant of ecosystem structure and function. The effects of soil temperature reverberate across diverse biotic and abiotic facets, delineating its pivotal role in shaping biogeochemical cycling, plant physiology, and broader environmental dynamics.

The pertinence of soil temperature emanates from its pivotal role in governing biogeochemical cycling. Soil temperature exerts a pronounced influence on microbial enzymatic kinetics, thereby regulating the rates of organic matter decomposition and nutrient mineralization. This fundamental process is recognized to underpin ecosystem nutrient availability and cycling dynamics, subsequently influencing primary productivity and trophic interactions. As noted by (Adair et al., 2008), microbial activity, intricately linked to temperature, orchestrates nutrient cycling and availability within soil ecosystems, thereby underscoring the central role of soil temperature in nutrient dynamics. Furthermore, soil temperature is an orchestrator of plant physiological responses, with implications for vegetation dynamics and distribution patterns. The thermal regime directly modulates phenological phases, such as germination, vegetative growth, and flowering, thereby shaping species establishment and competitive interactions within plant communities. Notably, research by (Bykova et al., 2012) elucidates how deviations in temperature regimes affect plant phenology, thereby disrupting synchrony within ecosystems and potentially leading to shifts in species composition. The effects of soil temperature traverse terrestrial domains and extend into aquatic ecosystems. Altered soil temperature regimes can propagate thermal fluctuations into adjacent water bodies, influencing aquatic organism physiology and distribution. This interconnectivity is notably exemplified by the study of (Fabris et al., 2019), elucidating how soil temperature-induced thermal changes influence stream temperatures, thus affecting aquatic species' metabolic rates and habitat suitability. Additionally, soil temperature is intricately intertwined with carbon dynamics and greenhouse gas emissions. Elevated soil temperatures accelerate microbial respiration rates, amplifying carbon dioxide releases into the atmosphere. This, in turn, aggravates the phenomenon of climate change and contributes to global carbon fluxes. Noteworthy in this context is the research by (Davidson and Janssens, 2006), wherein they elucidate the intricate feedback mechanisms between soil temperature, microbial activity, and greenhouse gas emissions.

In light of the aforementioned substantial impacts exerted by soil temperature on the entirety of the ecosystem, the precise determination of soil temperature has emerged as a prevalent and pivotal subject of inquiry. Soil temperature serves as a linchpin in biogeochemical cycles, orchestrating microbial activities that drive nutrient cycling and ecosystem functioning. To fathom nutrient dynamics, it is imperative to discern how soil temperature interacts with parameters such as soil moisture, organic matter content, and microbial community composition. The accurate measurement of soil temperature elucidates these intricate relationships, enabling researchers to predict nutrient release rates, optimize nutrient management strategies, and ensure the resilience of ecosystems. In the realm of plant ecology, soil temperature is a decisive arbiter of phenological transitions and species distributions. Accurate estimation of soil temperature offers insights into its interplay with factors like solar radiation, moisture availability, and vegetation cover. These insights empower ecologists to anticipate shifts in plant life strategies, aiding in the formulation of conservation tactics that ensure species persistence, preserve genetic diversity, and restore degraded landscapes. Furthermore, the thermal connectivity between terrestrial and aquatic realms hinges upon precise soil temperature estimation. By deciphering how soil temperature fluctuations propagate into adjacent water bodies, scientists gain a holistic understanding of cross-ecosystem thermal interactions. This knowledge is essential for safeguarding aquatic habitats, projecting species' responses to thermal variations, and implementing measures that preserve the intricate web of trophic relationships that characterize aquatic food webs. In the broader context of climate change, soil temperature estimation is an indispensable tool for gauging the impact of temperature alterations on carbon dynamics. Accurate measurements elucidate the nexus between soil temperature and microbial activity, a cornerstone of carbon cycling. As global efforts focus on curbing greenhouse gas emissions, soil temperature estimation empowers us to monitor carbon fluxes, optimize land management strategies, and contribute to a sustainable carbon-neutral future.

Soil temperature, a fundamental environmental parameter, depends on a range of intricate factors that collectively influence its variability. These factors include seasonality, solar radiation, air temperature, soil moisture, soil composition, and vegetation cover. In the light of information based on these factors, soil temperature can be measured using data provided from resources such as thermocouples, resistance thermometers, soil temperature probes, infrared thermometers, satellite-based sensors, and so on. On the other hand, machine learning-based methods have become very popular in recent years for soil temperature prediction, offering a data-driven approach to unravel the complexity of soil temperature dynamics, enabling informed decision-making, contributing to a deeper understanding of environmental systems and automating all these processes.

In this study, a pioneering methodology is introduced employing transformer models for the explicit purpose of prognosticating soil temperature. Within the scope of current knowledge, the utilization of transformer models within this framework constitutes an inaugural endeavor aimed at prognosticating soil temperature. Empirical investigations are conducted encompassing six distinct FLUXNET stations, encompassing their encapsulation through five distinct transformer models: Vanilla Transformer, Informer, Autoformer, Reformer, and ETSformer. To underscore the efficacy of the proffered framework, empirical findings are juxtaposed against both deep learning methodologies and extant scholarly investigations. The empirical findings not only corroborate the substantive value of employing transformer models but also ascertain their pivotal role in augmenting the landscape of scholarly contributions, thus establishing a seminal paradigm shift in the domain. The primary contributions of this work encompass the following aspects:

· A novel transformer-based model to improve the predictive performance of soil temperature.

· The first attempt to model time series data for the purpose of determining soil temperature using transformer methods

· Extensive experimental evaluation of proposed model against different transformer models and deep learning techniques to demonstrate the superior predictive performance of the proposed model.

· The evaluation of an efficient model with generalization capability sensitive to diverse station information and time steps.

· Superior prediction performance compared to the state-of-art studies.

This paper is organized as follows: Section 2 presents the literature on soil temperature estimation and related topics. Section 3 introduces the data collection stages, methods used in the study, and the proposed methodology. Experimental setup and results are given in Section 4. Section 5 summarizes and discusses results and outlines future research directions.

## 2. Related Work

This chapter provides a summary of literature studies centering on soil temperature prediction. In addition to studies focusing on soil temperature prediction, we also review studies where soil temperature is used or mentioned in many areas such as soil moisture and irrigation management, plant growth and productivity, seed germination and seedling development, soil organisms and microorganisms, soil chemistry and nutrients, diseases and pests, soil erosion and surface temperature, environmental changes and climate impacts.

A total of 129 scholarly articles centered around the utilization of deep learning (DL) applications in the domain of agriculture have been thoroughly examined (Attri et al., 2023), with the intention of classifying them into five distinct domains of application: namely, crop yield prognosis, plant stress identification, identification of weeds and pests, malady detection, and the domain of intelligent agricultural practices. This umbrella term of "smart farming" has been further dissected into subcategories encompassing water resource management, seed examination, and the comprehensive analysis of soil composition. The current scholarly inquiry underscores the intrinsic promise embedded within deep learning methodologies, discernibly enhancing agricultural output and invariably fostering economic advancement. The inquiry establishes that supervised learning networks, including Convolutional Neural Networks (CNN), Recurrent Neural Networks (RNN), AlexNet, and ResNet, are notably preeminent within the agricultural sector, instrumental in augmenting economic prosperity. This study defends that the imperativeness of persistent scholarly exploration in this sphere, in order to fully harness the potential inherent within deep learning for the facilitation of intelligent agricultural practices, ultimately culminating in the realization of a sustainable agricultural milieu.

A systematic literature review is conducted in the study (Khan et al., 2022a), focusing on hyperspectral imaging technology and the advanced deep learning and machine learning algorithms that are employed in agricultural applications, with the intention of extracting and amalgamating significant datasets and algorithms. The study reveals that the Hyperion hyperspectral, Landsat-8, and Sentinel 2 multispectral datasets have predominantly been utilized in the context of agricultural applications. The machine learning method that has witnessed the highest frequency of application is the support vector machine and random forest algorithms. Furthermore, the crop classification task is primarily facilitated by the deep learning-based Convolutional Neural Networks (CNN) model, a choice driven by its proven efficacy when handling hyperspectral datasets. It is anticipated that the insights curated within the purview of this present review shall be of substantial utility to nascent researchers engrossed in the hyperspectral remote sensing sphere, particularly within the context of agricultural applications, wherein machine and deep learning methodologies are pertinent.

In the study (Zhang et al., 2023), the response of soil respiration (Rs) to temperature and precipitation treatments within terrestrial ecosystems is examined by amalgamating studies encompassing 178 temperature treatments and 134 precipitation treatments. The results demonstrate that average increments in warming and precipitation respectively augment Rs by 13.1% and 33.1%. Additionally, the authors report that the positivity of the effect sizes pertaining to Rs is observable across diverse global variables such as mean annual temperature (MAT), mean annual precipitation (MAP), elevation, and the duration of experimentation (DUR). It is emphasized that this discernible trend signifies a robust reliance of Rs upon the broader spectrum of global climatic conditions. Furthermore, it is expressed that the interaction between other environmental factors and precipitation and temperature is striking, inducing an indirect alteration upon Rs.

For hourly soil temperature estimation, (Seifi et al., 2021) offer a hybrid model. Sunflower optimization (SFO), Firefly algorithm (FFA), Salp swarm algorithm (SSA), and particle swarm optimization (PSO) optimization algorithms are combined with adaptive neuron fuzzy interface system (ANFIS), support vector machine (SVM), radial basis function neural network (RBFNN), and multilayer perceptron (MLP) machine learning algorithms for this purpose to obtain better prediction accuracy of soil temperature. In Iran, experiments are conducted hourly at soil depths of 5, 10, and 30 cm. According to authors, when hybrid meta-heuristic models are used, the suggested technique yields results for soil temperature prediction that are trustworthy and accurate. (Singhal et al., 2021) employ an ANN model for the estimation of soil temperature of Central Himalaya region. Three layered nine different ANN models are constructed to forecast the temperatures of multi-depth soil. To show the best model of ANN variations, experiments proceeded on concurrent and preceding air-soil temperature data. The paper is concluded that the proposed ANN model ensures robust prediction results for the glacial fore-field regions of the Central Himalaya. (Bayatvarkeshi et al., 2021) focus on predicting soil temperature involving air temperature attributes in various climatic circumstances with the aid of machine learning algorithms. For this purpose, ANN and co-active neuro-fuzzy inference systems (CANFIS) models blended with wavelet transformation and called WANN and WCANFIS, respectively. To demonstrate the effectiveness of the proposed approach, experiments are implemented the data gathered at 12 location in Iran at 5, 10, 20, 30, 50 and 100 cm soil depths over the period 2000–2010 years. Experimental results indicate that using the hybrid WCANFIS model to estimate soil temperature has a better predictive capability, with a 0.43 RMSE score.

(Malik et al., 2022) presents a hybrid model combining machine learning algorithms with optimization techniques for the purpose of estimating daily soil temperature in a semi-arid zone of India. SVM, MLP, and ANFIS techniques are employed blending them with slime mould algorithm (SMA), particle swarm optimization (PSO), and spotted hyena optimizer (SHO) algorithms. Consolidating SVM and SMA models demonstrates the best MAE scores at 5-cm, 15-cm, and 30-cm soil depths. (Guleryuz, 2022) proposes to forecast the soil temperature of Giresun and Bayburt stations in Turkey employing machine learning algorithms. Bayesian Tuned Gaussian Process Regression (BT-GPR), Bayesian Tuned Support Vector Regression (BT-SVR), and Long-Short Term Memory (LSTM) techniques are modeled on a five and half year daily data. Author concludes the study that BT-GPR approach exhibits the best predictive performance with 0.0525 of RMSE score for estimating soil temperature at a 5-cm depth. (Li et al., 2022) develop an attention-aware long short-term memory network model to forecast soil moisture and temperature. Experiments are carried out on 1 and 7 daily FLUXNET dataset by comparing the proposed model with random forest (RF), support vector regression (SVR), elastic-net (ENET), original LSTM techniques. Experiment results show that the proposed LSTM model mostly exhibits superior prediction performance when compared to the literature studies. (Wang et al., 2023) introduce 1D-CNN-MLP neural network framework for predicting soil temperature in Yangling station of China between 2018 and 2021 years. In order to demonstrate the efficacy of the proposed model, a comparative analysis is conducted with respect to the performance of the MLP and LSTM models. Based on the obtained experimental outcomes, the authors reach a definitive conclusion that the utilization of the 1D-CNN-MLP model yields superior predictive accuracy compared to alternative techniques. This finding underscores the enhanced efficacy and potential of the proposed model in addressing the challenges of predictive modeling within the specific domain under investigation.

(Orhan et al.) present prediction of soil temperature employing LSTM memory networks to demonstrate the efficiency of the deep learning approach. The authors focus on detecting the soil temperature in the Bingöl zone during the period of 2013-2021 at 50 cm soil depth. Experiment results show that the proposed LSTM model achieves a 1.25 RMSE score. (Kucuk et al., 2022) propose a novel machine learning-based framework which provides an ordinal categorization of soil temperature. To obtain the best-fitted algorithm, five different conventional machine learning algorithms, namely decision tree, naive Bayes, k-nearest neighbors, support vector machines, and random forest are carried out. The dataset is gathered from the three states of the United States of America at 2, 4, 8, 20, and 40 inches of soil depths during the period of 2011-2020. The study concluded that the decision tree model exhibits the best classification performance ensuring a 90.95% accuracy score. In the context of spatial interpolation of soil temperature and water content within the land-water interface of southeast Canada, (Imanian et al., 2023) propose a methodology that harnesses the power of radial basis function neural networks and deep learning techniques. Through empirical experimentation, the authors demonstrate the impressive capabilities of these methods in effectively interpolating soil temperature data, even in regions characterized by abrupt transitions. The results highlight the potential of radial basis function neural networks and deep learning approaches in addressing the complex challenges associated with spatial interpolation in geographically intricate domains. (Bilgili et al., 2023) concentrate on the performance of machine learning algorithms for soil temperature estimation in the Sivas region at 5 cm, 50 cm, and 100 cm soil depths. Comprehensive experiments are carried out using the ANFIS network with fuzzy c-means (ANFIS-FCM), grid partition (ANFIS-GP), subtractive clustering (ANFIS-SC), feed-forward neural network (FNN), Elman neural network (ENN), and LSTM neural network methods. Authors report that the best prediction performances are ensured by ENN at 5 cm soil depth, FNN at 50 cm soil depth, and ANFIS-GP at 100 cm soil depth.

(Tüysüzoğlu et al., 2022) introduce a new model called self-training for the purpose of predicting soil temperature in the Izmir region. The hourly customized soil data is gathered by IoT devices and meteorological data is gathered for roughly nine months. The authors inform that the proposed technique when combined with the XGBoost technique performs the best success ranging from 0.385 to 2.888 MSE scores for all soil depths (10, 20, 30, 40, and 50 cm). (Imanian et al., 2022) perform an extensive comparison of thirteen approaches from conventional machine learning algorithms to advanced AI techniques for estimating soil temperature in the Ottowa region. Experiment results show that the deep learning approach exhibits the best prediction performance with an R-squared of 0.980 and NRMSE of 2.237%. In the investigation conducted by (Khan et al., 2022b), a novel approach based on ANN is presented for the estimation of soil temperature in Yazoo clay. The study focuses on six instrumented slopes located within a 25-mile radius of metropolitan Jackson, Mississippi. Over a period of seventeen months, comprehensive data on volumetric moisture content, precipitation, air temperature, and soil temperature at a depth of 1.5 meters (5 feet) at the crest of the slopes is collected. Notably, the adoption of the Levenberg-Marquardt (LM) algorithm in conjunction with the Tan-sigmoid transfer function yields remarkable results, showcasing the efficacy of the proposed method in accurately predicting soil temperature. These findings underscore the potential of ANN-based approaches in enhancing our understanding of soil temperature dynamics and aiding in environmental assessments. (Tuysuzoglu et al., 2022) propose a novel approach for predicting multi-view multi-depth soil temperature utilizing machine learning and time series techniques. The proposed framework is modeled by consolidating antecedent soil data and past meteorological data. Experiment results indicate that utilization of the support vector regression technique exhibits superior performance compared to the other methods in terms of approaching forecasted values of soil temperature to the actual values. (Wang et al., 2021) design a novel embedded approach for predicting soil temperature in Switzerland. To implement the proposed framework, a gated recurrent unit (GRU) model is constructed to gather the global and local attributes of the soil temperature data. The data collected from Laegern and Fluehli stations is composed of 5 cm, 10 cm, and 15 cm soil depths at 6hrs, 12hrs, and 24hrs time points. In addition to the GRU model, ANN, extreme learning machine model (ELM), and LSTM methods are modeled to demonstrate the contribution of the proposed framework. The authors conclude the study that the utilization of the GRU model achieves the best predictive performance with the minimum RMSE, MAE, and MSE scores. (Abimbola et al., 2021) introduce a knowledge-aided machine learning approach for boosting the daily prediction of soil temperature across the United States. With the usage of the ANFIS method, a new transformation of meteorological parameters is employed in order to boost the accuracy performance in soil temperature by varying soil depths as 5, 10, 20, 50, and 100 cm. To forecast the soil temperature of each soil depth, nontransformation (NT), autocorrelation (AC), moving average (MA), and a combination of transformations (NT-ACMA) of meteorological features are employed. Experiment results show that the success of the NT-ACMA model generally exhibits the best accuracy score with 0.99 of R-squared in both soil textures.

Within the scope of this research, we present an innovative framework that harnesses the capabilities of transformer models, setting it apart from prior studies in the literature. Notably, to the best of our knowledge, this study represents the inaugural exploration of transformer models in the scope of soil temperature forecasting. To demonstrate the contribution of the proposed model, experiments are carried out on six different FLUXNET stations namely Netherlands, France, Belgium, Italy, Finland and Switzerland. Furthermore, the experiment results are compared with standalone deep learning techniques and literature studies to ensure a fair comparison. Experiment results indicate that the usage of transformer models exhibits considerable prediction performance, therefrom establishing the new state-of-the-art.

## 3. Proposed Framework

Designed for NLP tasks, transformer models are also popular in computer vision, time series forecasting, and various other topics. One main reason for this popularity is that transformer models can capture long-time dependencies in sequential data. In this work, we compare the deep learning and transformer-based models to predict soil temperature in different time steps. Based on the results, the transformer-based models achieve state-of-the-art results for forecasting multivariate multistep time series data. The overall architecture of the proposed method is given in Figure 1. In Figure 1, there are two separate ways to forecast soil temperature. A performance comparison is conducted by estimating soil temperature using traditional LSTM and CNN-based methods in contrast to the proposed transformer-based techniques.

### 3.1. FLUXNET: Pre-processing

FLUXNET (Pastorello et al., 2020), is a dataset that includes observations of atmospheric variables in various parts of the world, as well as the exchange of energy, water and carbon dioxide between the land surface and the atmosphere. It is a comprehensive global dataset containing data from more than 500 tower sites in more than 50 countries. Data from FLUXNET can be used to study a wide variety of issues related to Earth's carbon and water cycles, including:

- The role of different types of ecosystems (such as forests, grasslands and farmlands) in the global carbon cycle.

- Factors affecting the exchange of water, energy and carbon dioxide between the land surface and the atmosphere.

- Impacts of land use and land cover change on carbon and water cycles.

- Responses of ecosystems to changing climatic conditions such as changes in temperature, precipitation and atmospheric CO2 concentrations.

Researchers, policymakers and other stakeholders can also use the data from Fluxnet2015 to better understand the processes driving the Earth's carbon and water cycles and inform decision-making on issues such as climate change, land use and ecosystem management.

The primary objective of the pre-processing stage revolves around the identification and removal of noisy data, which has the potential to significantly impede the performance of deep learning models. This crucial step involves the implementation of conditions designed to filter out outlier values, ensuring the integrity and reliability of the data. Subsequently, the time-series data is analyzed to identify and observe key components such as trend and seasonality. Within the scope of this study, our focus lies in constructing deep learning models and transformer-based models utilizing data obtained from six distinct FLUXNET sites. These sites are shown in Figure 2. By selecting these specific sites, a comprehensive analysis becomes possible. These sites encompass diverse geographical regions, enabling a thorough evaluation of the proposed models. This approach contributes to the robustness of the study, as it allows for a broader examination of the models' performance across different environmental conditions and characteristics.

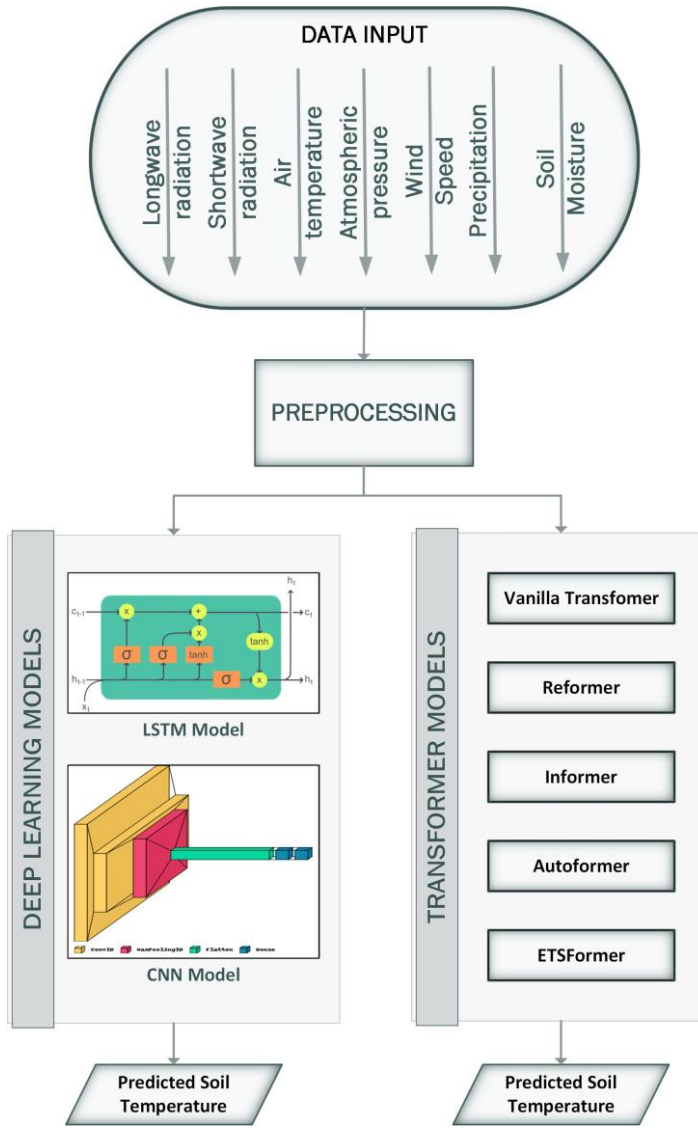

**Figure 1:** The flowchart of the proposed model.

The FLUXNET dataset contains data for each station up to 2014. Researchers have shared this dataset in various temporal resolutions, including hourly, daily, weekly, monthly, and yearly intervals. In this study, time series forecasting is carried out on hourly data. This hourly time series forecasting approach allows for a more detailed and fine-grained examination of the data's behavior and fluctuations. The proposed deep learning and transformer-based methods utilize the features listed in Table 1 as inputs. In Table 1, these features are coupled with their corresponding statistical outcomes, which are specific to the individual sites being analyzed and their metric units.

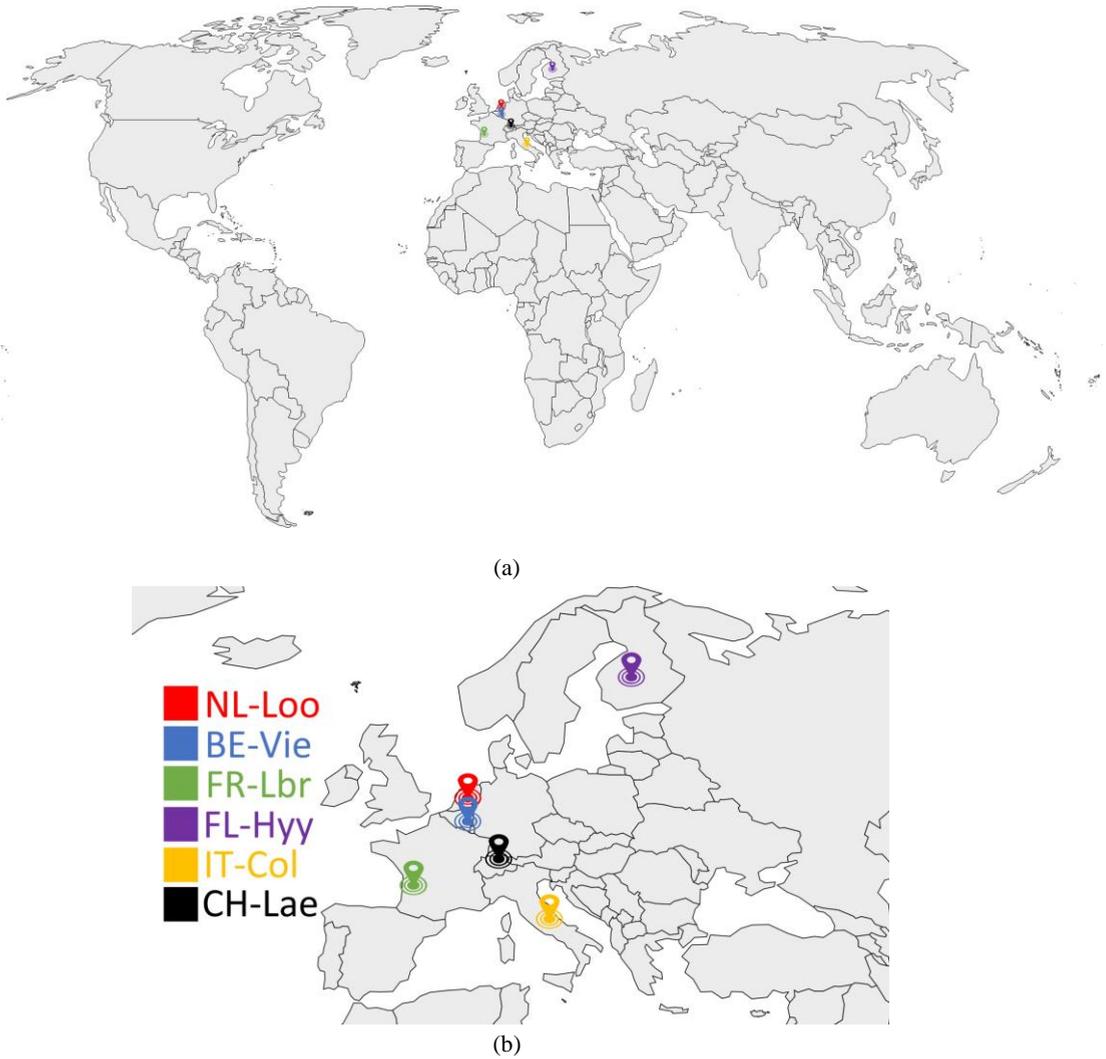

**Figure 2:** FLUXNET data sites used in this work. (a) On World map (b) On Zoomed World map

**Table 1**
Statistical results of features

| Features | NL - Loo | | | FR - Lbr | | | BE - Vie | | | IT - Col | | | FI - Hyy | | | CH-Lae | | |
|---|---|---|---|---|---|---|---|---|---|---|---|---|---|---|---|---|---|---|
| | Min | Max | Mean | Min | Max | Mean | Min | Max | Mean | Min | Max | Mean | Min | Max | Mean | Min | Max | Mean |
| Longwave radiation (W/m2) | 188.7 | 454.083 | 337.726 | 192.81 | 457.64 | 334.129 | -9999 | -9999 | -9999 | 142.891 | 388.976 | 281.703 | 152.786 | 432.23 | 301.933 | 135.786 | 423.858 | 304.537 |
| Shortwave radiation (W/m2) | 0 | 1005 | 116.155 | 0 | 1017.46 | 146.771 | 0 | 1012.67 | 114.1476 | 0 | 1155.14 | 178.35 | 0 | 855.4 | 91.237 | 0 | 1074.41 | 136.899 |
| Air temperature (deg C) | -14.86 | 34.82 | 10.103 | -7.71 | 36.18 | 12.877 | -15.31 | 34.23 | 8.3922 | -14.055 | 28.601 | 7.249 | -26.57 | 32.16 | 4.056 | -17.12 | 31.82 | 7.784 |
| Atmospheric pressure (kpa) | 96.053 | 104.169 | 101.084 | 98.198 | 103.85 | 101.573 | 91.505 | 98.546 | 96.0348 | 82.119 | 86.631 | 84.895 | 94.581 | 103.374 | 99.148 | 89.617 | 95.295 | 93.244 |
| Wind speed (m/s) | 0.024 | 9.377 | 2.352 | 0.039 | 12.813 | 3.114 | 0.013 | 10.095 | 2.442 | 0.025 | 6.94 | 1.625 | 0.04 | 9.838 | 3.27 | 0.004 | 10.497 | 2.23 |
| Precipitation (mm) | 0 | 7.641 | 0.047 | 0 | 1.709 | 0.039 | 0 | 5.471 | 0.0725 | 0 | 5.943 | 0.098 | 0 | 3.008 | 0.047 | 0 | 3.551 | 0.0675 |
| Soil moisture (% by volume) | 0 | 25.1 | 8.6 | 17.711 | 94.447 | 38.396 | 14.7 | 43.8 | 31.271 | 11.324 | 57.785 | 31.511 | 5.407 | 64.907 | 27.366 | 7.7 | 32.74 | 21.89 |

## 3.2. Forecasting

This work focuses on forecasting soil temperature by employing both deep learning-based models and transformer-based models. The aim is to develop predictive methodologies that utilize these two distinct approaches to accurately anticipate soil temperature values over time. By comparing the performance of deep learning and transformer models in this context, the study aims to determine which approach yields more effective and precise predictions for soil temperature fluctuations.

### 3.2.1. Deep learning methods

Today, **Convolutional Neural Networks (CNNs)** are widely used in image and video processing applications. CNNs, which can be considered as a type of deeper layered neural networks, stand out with their feature extraction capabilities far beyond the features used in traditional machine learning approaches.

The motivation behind the development of CNNs was to create a more efficient approach to image processing, by exploiting the local correlation between neighboring pixels. Unlike traditional Artificial Neural Networks (ANNs), which treat all input features as independent, CNNs apply convolutional filters to the input data, resulting in more efficient use of the network's parameters.

In terms of architectural distinctions, CNNs diverge from simple ANNs by incorporating convolutional layers in lieu of fully connected layers. Convolutional layers play a key role in CNNs as they facilitate the application of learned filters to the input data. These filters, also referred to as kernels or weights, are compact matrices that traverse the input data through a sliding window mechanism. Through element-wise multiplication and subsequent summation operations, the filters generate scalar outputs. After that, a crucial non-linearity is added to the model using a non-linear activation function, such as the Rectified Linear Unit (ReLU). The proposed CNN-based methodology is visually illustrated in Figure 3.

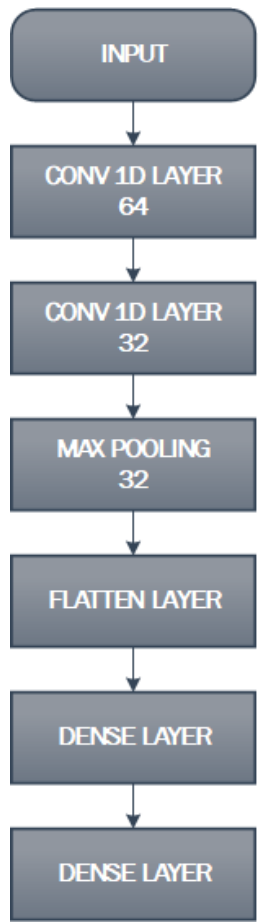

**Figure 3:** The architecture of the proposed CNN model

As a type of recurrent neural networks (RNN), **Long-short Term Memory (LSTM)**, stands as a powerful tool tailored to tackle the intricacies of sequential data, encompassing domains like time series analysis, natural language processing, and audio processing. Devised by Hochreiter and Schmidhuber in 1997, LSTM networks swiftly emerged as a favored option across diverse fields of application. The goal of LSTM networks is to solve the problem of typical

RNNs' vanishing gradients, which might make it challenging for the network to learn long-term relationships. With the use of a novel sort of memory cell that can selectively store or remove data over time, LSTM networks enable the network to keep track of pertinent data from earlier time steps.

The main difference between LSTM networks and other types of neural networks, such as simple ANNs and CNNs, is that LSTM networks have a recurrent structure that allows them to process sequential data. LSTM networks revolutionize information processing through a pioneering mechanism centered on a memory cell governed by three pivotal gates: the input gate, the forget gate, and the output gate. These gatekeepers skillfully manage the ebb and flow of data, adeptly determining which fragments to embrace or discard within the memory cell's sanctuary. The result is a seamless fusion of selective information storage and erasure, affording the network unparalleled adaptability. During training, the LSTM network learns to adjust the parameters of the input, forget, and output gates, as well as the candidate memory cell value, to selectively store or erase information over time. As a result, the network is able to keep track of important data from earlier time steps and utilize that memory to forecast time steps in the future. The proposed method LSTM-based method is presented in Figure 4.

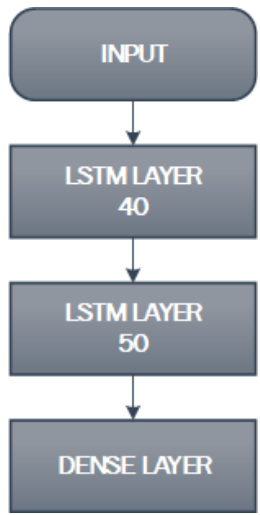

**Figure 4:** The architecture of the proposed LSTM model

### 3.2.2. Vanilla transformer

**The concept of transformers,** as presented in the seminal work by Vaswani et al. in 2017 (Vaswani et al., 2017), encompasses a profound deep-learning architecture. Specifically designed for the analysis of sequential data, transformers exhibit exceptional proficiency in domains such as natural language and time series. At their core, transformers consist of two essential components: an encoder and a decoder. While the encoder diligently handles the processing of the input sequence, the decoder takes charge of generating the corresponding output sequence. Notably, the transformative aspect lies in the self-attention mechanism, a groundbreaking innovation that endows the model with the remarkable ability to selectively focus on distinct segments of the input sequence during element-wise processing. As a result, transformers excel in capturing long-range dependencies within the input sequence, surpassing the capabilities of conventional recurrent neural networks.

$$\text{Attention}(Q, K, V) = \text{softmax}\left(\frac{QK^T}{\sqrt{d_k}}\right)V \tag{1}$$

The scaled-dot product attention proposed in the article is given in (1). This attention mechanism is then stacked, forming Multi-Head attention. One of the drawbacks of the scaled-dot product attention is that it has the computational complexity of $O^2$, where L is the length of the sequence which makes the transformer models computationally expensive to train.

We train the vanilla transformer model on six different fluxnet site datasets using default parameters suggested in the original paper for different time steps.

### 3.2.3. Reformer: The Eficient Transformer Model

Large Transformer models achieve state-of-the-art results across a range of tasks, but training these models can be very costly, especially if to be considered in long sequences. Authors of the Reformer model (Kitaev et al., 2020) introduce two new methods to improve transformer model efficiency. They first replace the scaled-dot product attention with locality-sensitive hashing (LSH), resulting decrease in computational complexity from $O^2$ to $O(\log O)$. The second main contribution is the use of reversible layers. It was first introduced by Gomez et al. (2017) (Gomez et al., 2017) ensures that only a single copy of the activations is stored in the entire model, thus eliminating the N factor.

Consider an instance wherein the model encompasses a word sequence of 64,000 lexical units. The LSH attention orders this sequence in a manner of similarity and calculates attention with the most similar keys, thus reducing complexity. The QKT matrix multiplication in (1) is the main reason for $\theta$ complexity. The first idea of the Reformer is instead of multiplying all of the query Q with the Keys K, indexing all queries individually that have the dimension of $1 \times d_k$, thus avoiding $O^2$ complexity. The LSH attention proposed in Reformer is given in (2).

$$\text{Attention}(Q, K, V) = \text{softmax}\left(\frac{q_i K^T}{\sqrt{d_k}}\right) V \tag{2}$$

RevNets, a novel approach to network architecture, introduce a distinctive mechanism that diverges from conventional methods of backpropagation. In the context of RevNets, layers are reversed incrementally, enabling backpropagation to traverse from the output to the input of the network without the need to access intermediate values during the backward pass. A key characteristic of RevNets lies in the operation of reversible layers, which operate on pairs of inputs and outputs, denoted as $x_1, x_2 \rightarrow y_1, y_2$ The behavior of these reversible layers can be accurately described using the following equations:

$$y_1 = x_1 + F(x_2) \quad y_2 = x_2 + G(y_1) \tag{3}$$

The attention and Feed-forward layers are combined by the authors to apply the RevNet concept to transformers. F becomes an attention layer in the notations in (3), whereas G becomes the feed-forward layer.

$$Y_1 = X_1 + \text{Attention}(X_2) \quad Y_2 = X_2 + \text{FeedForward}(Y_1) \tag{4}$$

### 3.2.4. Informer

The Informer model presented in (Zhou et al., 2021) proposes a new transformer-based architecture called "Informer" for time-series forecasting that can handle long sequences efficiently. Long sequence time-series forecasting (LSTF) tasks cannot be performed using Transformer models due to a number of problems. The proposed Informer model makes a number of significant additions to the usage of transformer-based models in LSTF tasks.

Informer introduces a hybrid attention mechanism that combines a ProbSparse Self-Attention (PSA) mechanism and a Local-Global (LG) structure to capture both local and global dependencies in time-series data. This mechanism enables the model to focus on important parts of the sequence while efficiently processing long sequences.

The proposed generative style decoder in Informer makes predictions for long time-series sequences in a single forward operation rather than to in a sequential manner. This considerably improves the speed of inference for long-sequence predictions. Each key can only attend to dominant queries with the help of ProbSparse Self-attention:

$$A(Q, K, V) = \text{Softmax}\left(\frac{\bar{Q} K^T}{\sqrt{d}}\right) V \tag{5}$$

The matrix Q in (5) has the same dimensions as q and is sparsely populated with only the Top-u queries based on the sparsity measurement M(q, K). With the ProbSparse attention, like Reformer, the Informer transformer model achieves $O(L \log L)$ complexity.

### 3.2.5. Autoformer

Time series forecasting finds extensive application in fields such as energy consumption, traffic and economics planning, weather, and disease propagation prediction. In real-world scenarios, there is an urgent need to forecast further into the future, which is essential for effective long-term planning and early warning systems. Consequently, the research paper focuses on studying the long-term forecasting problem of time series, which is defined by the prediction of an extended time series with a substantial length.

The forecasting task becomes exceedingly difficult when dealing with long-term settings. Firstly, identifying temporal dependencies directly from extended time series is unreliable as these dependencies may be concealed by intertwined temporal patterns. Secondly, conventional Transformer models with self-attention mechanisms become computationally expensive for long-term forecasting due to the quadratic complexity of sequence length.

The researchers stated that while previous transformer-based forecasting models (Kitaev et al., 2020; Zhou et al., 2021) decreased the computational cost by presenting sparse self-attention, these models still utilize point-wise representation aggregation. While aiming to enhance efficiency, the utilization of information may be compromised due to the presence of sparse point-wise connections.

Drawing from the issues mentioned above, the researchers present a novel solution called Autoformer for long-term time series forecasting, which replaces the conventional Transformers. While retaining the residual and encoder-decoder structure, the Autoformer adopts a decomposition forecasting architecture in contrast to the Transformer. Autoformer draws inspiration from the stochastic process theory (Chatfield, 2003; Papoulis and Unnikrishna Pillai, 2002) and incorporates an Auto-Correlation mechanism instead of the self-attention mechanism. This mechanism leverages series periodicity to detect sub-series similarity and aggregates related sub-series from underlying periods. By adopting this series-wise mechanism, Autoformer reduces the computational complexity to $O(L \log L)$ for length-L series and eliminates the information consumption bottleneck by extending point-wise representation aggregation to the level of sub-series. On six metrics, Autoformer obtains the highest level of accuracy.

### 3.2.6. ETSFormer

**(Woo et al., 2022)** created a novel time-series forecasting model known as ETSformer that combines two frameworks to enhance its performance. The ETSformer introduces novel exponential smoothing and frequency attention techniques coupled with the classical understanding of seasonal-trend decomposition and exponential smoothing with existing transformers. The inspiration behind ETSformer stems from classical exponential smoothing techniques employed in time-series forecasting. In place of the self-attention mechanism in the original Transformers, it uses cutting-edge exponential smoothing attention (ESA) and frequency attention (FA) methods, which improve both accuracy and efficiency. As a result, the ETSformer achieves state-of-the-art performance in time-series forecasting.

The researchers used Holt-Winters' method (Holt, 2004) for exponential smoothing which can be formulated as:

$$
\begin{aligned}
Level : \ & e_t = \alpha \left( x_t - s_{t-p} \right) + (1 - \alpha) \left( e_{t-1} + b_{t-1} \right) \\
Growth : \ & b_t = \beta \left( e_t - e_{t-1} \right) + (1 - \beta) b_t \\
Seasonal : \ & s_t = \gamma \left( x_t - e_{t-1} \right) + (1 - \gamma) s_{t-p} \\
Forecasting : \ & \hat{x}_{h|t} = e_t + h b_t + s_{t+h-p}
\end{aligned}
\tag{6}
$$

The equations mentioned in (6) represent the h-steps ahead forecast, where p represents the period of seasonality and $\hat{x}_{h|t}$ is the forecast. These equations show that the forecast is made up of the last estimated level $e_t$, which is incremented by $h$ times the last growth factor, $b_t$, and the last available seasonal factor $s_{t+h-p}$ is added.

Within each layer of the proposed architecture, the encoder component is purposefully constructed to iteratively extract latent components associated with growth and seasonality from the provided lookback window. Similarly, the level component is extracted, employing a technique akin to classical level smoothing as represented by Equation (6). These extracted components are subsequently propagated to the decoder module, where they converge to generate the ultimate H-step ahead forecast. This forecast is synthesized by combining individual forecasts for level, growth, and seasonality, as precisely defined in the following expressions:

$$
\hat{X}_{t:t+H} = E_{t:t+H} + \text{Linear} \left( \sum_{n=1} \left( B_{t:t+H}^{(n)} + S_{t:t+}^{(n)} \right) \right)
\tag{7}
$$

**Table**
Dataset division and statistical information for six flux sites.

| Station | Number of Observations (hourly) | Training Data Scope (70%) | Validation Data Scope (10%) | Test Data Scope (20%) |
|---------|--------------------------------|---------------------------|------------------------------|------------------------|
| NL - Loo | 177551 | 2000-01-01 2010-07-02 | 2010-07-02 2012-01-01 | 2012-01-01 2014-12-31 |
| BE - Vie | 131495 | 2000-01-01 2010-07-02 | 2010-07-02 2012-01-01 | 2012-01-01 2014-12-31 |
| FR - Lbr | 35063 | 2005-01-01 2007-10-20 | 2007-10-20 2008-03-14 | 2008-03-14 2008-12-31 |
| FI - Hyy | 46055 | 2009-09-30 2013-06-04 | 2013-06-04 2013-12-13 | 2013-12-13 2014-12-31 |
| IT - Col | 43841 | 2009-12-30 2013-07-01 | 2013-07-01 2013-12-31 | 2013-12-13 2014-12-31 |
| CH - Lae | 81851 | 2005-08-30 2012-03-13 | 2012-03-13 2013-02-17 | 2013-02-17 2014-12-31 |

The matrices $E_{t;t+H} \in \mathbb{R}^{H \times m}$ and $B^{(n)}_{t;t+H}$, $S^{(n)}_{t;t+H} \in \mathbb{R}^{H \times d}$ in the equation above stand in for the level predictions as well as the growth and seasonal latent representations of each time step in the forecast horizon. There are a total of N encoder stacks, and the index "n" denotes the stack index. In order to project the recovered growth and seasonal representations from latent to observation space, Linear(.): $R^d \to R^m$ must conduct an element-wise operation along each time step.

## 4. Experimental Results

This study employs the widely recognized FLUXNET dataset, containing soil temperature values, for the purpose of comparing and evaluating the proposed methods. In the experiments, time series forecasting is carried out on hourly data. The dataset division and statistical information for each six FLUXNET stations are given in Table 2. The performance of the proposed methods is measured using the well-known metrics given in (1) and (2). MAE (Mean Absolute Error), also known as L1 loss, is the average absolute error between actual and predicted values. MSE (Mean Squared Error), also known as L2 loss, is the average squared error between actual and predicted values. The utilization of both MAE and MSE computations offers insight into model performance across the entire dataset. The main distinction between squared error and absolute error is that squared error penalizes significant errors more severely than absolute error since it squares the errors rather than just computing the difference. This differentiation in treatment enhances the sensitivity of squared error to larger deviations.

$$MSE = \frac{1}{n} \sum_{i=1}^{n} (T_{real} - T_{pre})^2 \tag{8}$$

$$MAE = \frac{1}{n} \sum_{i=1}^{n} |T_{real} - T_{pre}| \tag{9}$$

Table 3 provides an overview of the training parameters utilized for both the deep learning-based models and the transformer-based models. These parameters remain consistent across all models of their respective categories, ensuring uniformity and comparability in the training process.

Evaluation results of deep learning-based methods, the details of which are explained in Section 2.2.1, are given in Table 4 and the evaluation results of the transformer-based methods, the details of which are given in 2.2.2 to 2.2.6, are given in Table 5. The results given in Table 4 and Table 5 are obtained using the distribution given in Table 2 for 6 different stations. Across these tables, each model is assessed for its predictive capability over various time intervals, including 96, 192, 336, and 720 time steps. These time steps signify the range into the future that the model forecasts. For instance, a time step of 96 indicates a prediction horizon of the subsequent 96 hours. Within Table 4 and Table 5, lower values indicate superior performance and are highlighted using bold fonts. The reformer method proved to be more successful than other methods in 11 out of 20 tests in 6 cities and 4-time steps. Transformers methods typically outperform deep learning methods in terms of results. Particularly noteworthy is the Reformer method's remarkable achievement, boasting results that are approximately 50 times more favorable compared to outcomes achieved using the CNN approach.

**Table 3**
Training parameters

| Parameters | Deep Learning Models | Transformer Models |
|---|---|---|
| Learning rate | 0.001 | 0.0001 |
| Epoch | 20 | 5 |
| Batch | 32 | 32 |
| Optimization algorithm | ADAM | GELU |

**Table 4**
Evaluation of conventional deep learning-based models

| Methods | Metrics | NL - Loo | | | | FR - Lbr | | | | BE - Vie | | | | IT - Col | | | | FI - Hyy | | | | CH - Lae | | | |
|---|---|---|---|---|---|---|---|---|---|---|---|---|---|---|---|---|---|---|---|---|---|---|---|---|---|
| | | 96 | 192 | 336 | 720 | 96 | 192 | 336 | 720 | 96 | 192 | 336 | 720 | 96 | 192 | 336 | 720 | 96 | 192 | 336 | 720 | 96 | 192 | 336 | 720 |
| LSTM | MSE | 4.006 | 4.685 | 20.459 | 5.636 | 4.030 | 4.116 | 5.850 | 2.388 | 3.015 | 3.571 | 3.892 | 4.517 | 4.656 | 6.338 | 7.171 | 2.152 | 2.345 | 3.364 | 3.247 | | 7.576 | 9.244 | 10.141 | 10.834 |
| | MAE | 1.517 | 1.68 | 3.944 | 1.887 | 1.519 | 1.565 | 1.682 | 1.933 | 1.215 | 1.322 | 1.45 | 1.543 | 1.495 | 1.542 | 1.898 | 1.997 | 1.0008 | 1.086 | 1.319 | 1.526 | 2.139 | 2.381 | 2.510 | 2.594 |
| CNN | MSE | **1.825** | 1.314 | 2.149 | 2.127 | 1.746 | 1.966 | 2.206 | 2.169 | 1.673 | 1.768 | 1.843 | 1.857 | 2.135 | 2.078 | 2.168 | 2.560 | 1.299 | 1.404 | 1.564 | 1.699 | 7.078 | 8.501 | 9.769 | 10.399 |
| | MAE | 1.342 | 1.464 | 1.646 | 1.673 | 1.321 | 1.496 | 1.741 | 1.699 | 1.285 | 1.374 | 1.425 | 1.448 | 1.563 | 1.616 | 1.767 | 2.026 | 0.955 | 0.985 | 1.079 | 1.225 | 2.143 | 2.280 | 2.489 | 2.575 |

**Table 5**
Evaluation of transformer-based models

| Methods | Metrics | NL - Loo | | | | FR - Lbr | | | | BE - Vie | | | | IT - Col | | | | FI - Hyy | | | | CH - Lae | | | |
|---|---|---|---|---|---|---|---|---|---|---|---|---|---|---|---|---|---|---|---|---|---|---|---|---|---|
| | | 96 | 192 | 336 | 720 | 96 | 192 | 336 | 720 | 96 | 192 | 336 | 720 | 96 | 192 | 336 | 720 | 96 | 192 | 336 | 720 | 96 | 192 | 336 | 720 |
| Transformer | MSE | 0.004 | 0.039 | **0.061** | 0.102 | **0.066** | 0.104 | 0.122 | 0.142 | 0.061 | 0.104 | 0.129 | 0.149 | **0.043** | 0.083 | 0.101 | 0.139 | 0.068 | 0.1 | 0.107 | 0.138 | **0.034** | 0.160 | 0.207 | 0.283 |
| | MAE | 0.107 | 0.142 | 0.188 | 0.23 | 0.203 | 0.25 | 0.288 | 0.327 | 0.184 | 0.25 | 0.274 | 0.312 | 0.143 | 0.199 | 0.233 | 0.249 | 0.187 | 0.236 | 0.247 | 0.273 | 0.262 | 0.305 | 0.343 | 0.364 |
| Informer | MSE | 0.02 | **0.036** | 0.062 | **0.09** | 0.068 | 0.098 | 0.132 | 0.172 | **0.059** | **0.104** | 0.123 | 0.152 | 0.047 | 0.092 | 0.119 | 0.113 | 0.071 | 0.097 | 0.111 | 0.129 | 0.121 | 0.148 | 0.187 | 0.217 |
| | MAE | 0.091 | 0.166 | 0.212 | 0.313 | 0.218 | 0.23 | 0.258 | 0.38 | 0.174 | 0.25 | 0.274 | 0.312 | 0.143 | 0.199 | 0.233 | 0.248 | 0.304 | 0.208 | 0.263 | 0.314 | 0.4 | 0.269 | 0.303 | 0.364 |
| Autoformer | MSE | **0.015** | 0.048 | 0.078 | 0.166 | 0.076 | 0.098 | 0.118 | 0.161 | 0.079 | 0.108 | 0.146 | 0.237 | 0.058 | 0.099 | 0.109 | 0.153 | 0.083 | 0.125 | 0.168 | 0.269 | 0.123 | 0.155 | 0.185 | 0.249 |
| | MAE | 0.091 | 0.166 | 0.212 | 0.313 | 0.23 | 0.259 | 0.283 | 0.347 | 0.199 | 0.248 | 0.298 | 0.398 | 0.189 | 0.234 | 0.271 | 0.337 | 0.195 | 0.23 | 0.276 | 0.385 | 0.259 | 0.309 | 0.342 | 0.412 |
| Reformer | MSE | 0.028 | 0.044 | 0.075 | 0.116 | 0.068 | 0.094 | **0.117** | **0.107** | 0.069 | 0.108 | **0.118** | **0.133** | 0.047 | **0.063** | **0.071** | **0.076** | 0.06 | 0.074 | 0.079 | **0.105** | 0.106 | **0.144** | **0.164** | **0.185** |
| | MAE | 0.128 | 0.158 | 0.213 | 0.269 | 0.208 | 0.251 | 0.279 | 0.269 | 0.201 | 0.258 | 0.274 | 0.295 | 0.145 | 0.174 | 0.189 | 0.206 | 0.179 | 0.204 | 0.213 | 0.248 | 0.251 | 0.297 | 0.321 | 0.344 |
| ETSformer | MSE | 0.023 | 0.049 | 0.085 | 0.189 | 0.086 | 0.111 | 0.131 | 0.193 | 0.069 | 0.106 | 0.149 | 0.257 | 0.062 | 0.097 | 0.128 | 0.186 | 0.071 | 0.1 | 0.144 | 0.275 | 0.116 | 0.160 | 0.191 | 0.276 |
| | MAE | 0.118 | 0.168 | 0.22 | 0.338 | 0.229 | 0.259 | 0.283 | 0.347 | 0.199 | 0.248 | 0.298 | 0.398 | 0.189 | 0.234 | 0.271 | 0.337 | 0.195 | 0.23 | 0.276 | 0.385 | 0.259 | 0.309 | 0.342 | 0.412 |

**Table 6**
Comparison of proposed method and literature

| Methods | Method Details | Time Step (hours) | NL - Loo | FR - Lbr | BE - Vie | IT - Col | FI - Hyy | CH-Lae |
|---|---|---|---|---|---|---|---|---|
| (Wang et al., 2021) | GRU & Auxiliary Network | 24 | - | - | - | - | - | 1.0287 |
| (Hao et al., 2021) | EEMD-CNN | 120 | - | - | - | - | - | 0.498 |
| (Li et al., 2022) | Attention-based LSTM | 168 | 0.352 | 0.811 | 1.119 | 2.947 | 3.268 | - |
| Proposed Method | Reformer | 192 | **0.158** | **0.251** | **0.258** | **0.174** | **0.204** | **0.297** |

Table 6 provides a comprehensive comparison between contemporary soil temperature prediction methods outlined in existing literature and the proposed approach, using the evaluation metric of Mean Absolute Error (MAE). The comparison is carried out on the Fluxnet. The compared methods in Table 6 are deep learning-based approaches and their method details are given in Table 6. The results of these methods are originally taken from their papers. The prediction values provided in Table 6 are derived for a soil depth of 5 cm. Since predicting short-term time steps is relatively less challenging, the longest prediction intervals are utilized for all methods in the table. It's important to point out that the results of the (Wang et al., 2021) and (Hao et al., 2021) methods are based on data from a limited number of stations. This implies that their performance evaluation is restricted to the conditions of these specific stations and might not accurately reflect their performance in broader scenarios. As seen in Table 6, the proposed Reformer method consistently demonstrates lower MAE values in comparison to the other methods. This consistent trend implies that the Reformer approach showcases superior predictive accuracy in forecasting soil temperature, positioning it as a noteworthy advancement in this domain. Moreover, it's worth highlighting that the Reformer method demonstrates outcomes that are notably superior by almost twice compared to the second-best method (Li et al., 2022), indicating its capacity to substantially mitigate prediction errors in contrast to existing approaches.

# 5. Discussion and Conclusion

Soil temperature is a crucial parameter that affects modifications among ecological stability, water and energy flows, and nutrient cycling both under and above the ground. It is known that there are many parameters such as solar radiation, air temperature, precipitation, wind speed, pressure gradient, and moisture that are dependent directly or implicitly on the soil temperature. In the lack of automatic predicting frameworks, the manual attempt for the prediction of soil temperature turns into a quite challenging task since modifications in the parameters alter remarkably over both long and short terms. The application of machine learning techniques in various sciences and engineering fields in recent years has gained an important place in the automatic prediction of soil temperature by eliminating this manual approach.

In this work, a transformer-based approach is proposed to predict soil temperature. In Earth Science, the predictions of soil temperature are crucial for understanding and forecasting planetary change. In the proposed method, firstly, deep learning-based CNN and LSTM methods are analyzed. Then, recently proposed transformer-based models are evaluated. The method with the highest performance obtained from the evaluation is also compared with the literature. The experimental results demonstrate that employing transformer-based models for predicting soil temperature using 1-dimensional signals and considering long-term time steps yields impressive outcomes. This underscores the efficacy of transformer-based methodologies in this context and underscores their potential for enhancing soil temperature prediction accuracy.

By accurately predicting soil temperature, through the proposed method, numerous benefits can be realized for ecology. This predictive capability enhances our comprehension of how temperature fluctuations impact ecosystems, species behavior, and habitat management. Furthermore, it aids in assessing the effects of climate change, optimizing resource allocation, and preserving biodiversity. The approach also contributes to understanding carbon cycling dynamics, informing conservation strategies, and providing early warnings for extreme temperature events. In essence, this innovation equips ecologists with valuable insights, enabling them to make informed decisions and gain a deeper understanding of the intricate relationships between soil temperature and ecological systems.

# References

Abimbola, O.P., Meyer, G.E., Mittelstet, A.R., Rudnick, D.R., Franz, T.E., 2021. Knowledge-guided machine learning for improving daily soil temperature prediction across the united states. Vadose Zone Journal 20, e20151.

Adair, E.C., Parton, W.J., Del Grosso, S.J., Silver, W.L., Harmon, M.E., Hall, S.A., Burke, I.C., Hart, S.C., 2008. Simple three-pool model accurately describes patterns of long-term litter decomposition in diverse climates. Global change biology 14, 2636–2660.

Attri, I., Awasthi, L.K., Sharma, T.P., Rathee, P., 2023. A review of deep learning techniques used in agriculture. Ecological Informatics , 102217.

Bayatvarkeshi, M., Bhagat, S.K., Mohammadi, K., Kisi, O., Farahani, M., Hasani, A., Deo, R., Yaseen, Z.M., 2021. Modeling soil temperature using air temperature features in diverse climatic conditions with complementary machine learning models. Computers and Electronics in Agriculture 185, 106158.

Bilgili, M., Şaban, Ü., ŞEKERTEKİN, A., GÜRLEK, C., 2023. Machine learning approaches for one-day ahead soil temperature forecasting. Journal of Agricultural Sciences 29, 221–238.

Bykova, O., Chuine, I., Morin, X., Higgins, S.I., 2012. Temperature dependence of the reproduction niche and its relevance for plant species distributions. Journal of Biogeography 39, 2191–2200.

Chatfield, C., 2003. The analysis of time series: an introduction. Chapman and hall/CRC.

Davidson, E.A., Janssens, I.A., 2006. Temperature sensitivity of soil carbon decomposition and feedbacks to climate change. Nature 440, 165–173.

Fabris, L., Buddendorf, W.B., Soulsby, C., 2019. Assessing the seasonal effect of flow regimes on availability of atlantic salmon fry habitat in an upland scottish stream. Science of The Total Environment 696, 133857.

Gomez, A.N., Ren, M., Urtasun, R., Grosse, R.B., 2017. The reversible residual network: Backpropagation without storing activations. Advances in neural information processing systems 30.

Guleryuz, D., 2022. Estimation of soil temperatures with machine learning algorithms—giresun and bayburt stations in turkey. Theoretical and Applied Climatology 147, 109–125.

Hao, H., Yu, F., Li, Q., 2021. Soil temperature prediction using convolutional neural network based on ensemble empirical mode decomposition. IEEE Access 9, 4084–4096.

Holt, C.C., 2004. Forecasting seasonals and trends by exponentially weighted moving averages. International journal of forecasting 20, 5–10.

Imanian, H., Hiedra Cobo, J., Payeur, P., Shirkhani, H., Mohammadian, A., 2022. A comprehensive study of artificial intelligence applications for soil temperature prediction in ordinary climate conditions and extremely hot events. Sustainability 14, 8065.

Imanian, H., Shirkhani, H., Mohammadian, A., Hiedra Cobo, J., Payeur, P., 2023. Spatial interpolation of soil temperature and water content in the land-water interface using artificial intelligence. Water 15, 473.

Khan, A., Vibhute, A.D., Mali, S., Patil, C., 2022a. A systematic review on hyperspectral imaging technology with a machine and deep learning methodology for agricultural applications. Ecological Informatics 69, 101678.

Khan, M.S., Ivoke, J., Nobahar, M., Amini, F., 2022b. Artificial neural network (ann) based soil temperature model of highly plastic clay. Geomechanics and Geoengineering 17, 1230–1246.

Kitaev, N., Kaiser, Ł., Levskaya, A., 2020. Reformer: The efficient transformer. arXiv preprint arXiv:2001.04451 .

Kucuk, C., Birant, D., TAŞER, P.Y., 2022. A novel machine learning approach: Soil temperature ordinal classification (stoc). Journal of Agricultural Sciences 28, 635–649.

Li, Q., Zhu, Y., Shangguan, W., Wang, X., Li, L., Yu, F., 2022. An attention-aware lstm model for soil moisture and soil temperature prediction. Geoderma 409, 115651.

Malik, A., Tikhamarine, Y., Sihag, P., Shahid, S., Jamei, M., Karbasi, M., 2022. Predicting daily soil temperature at multiple depths using hybrid machine learning models for a semi-arid region in punjab, india. Environmental Science and Pollution Research 29, 71270–71289.

Orhan, İ., Özkan, İ., ÖZTAŞ, T., YUKSEL, A., . Soil temperature prediction with long short term memory (lstm). Türk Tarım ve Doğa Bilimleri Dergisi 9, 779–785.

Papoulis, A., Unnikrishna Pillai, S., 2002. Probability, random variables and stochastic processes.

Pastorello, G., Trotta, C., Canfora, E., Chu, H., Christianson, D., Cheah, Y.W., Poindexter, C., Chen, J., Elbashandy, A., Humphrey, M., Isaac, P., Polidori, D., Ribeca, A., van Ingen, C., Zhang, L., Amiro, B., Ammann, C., Arain, M.A., Ardö, J., Arkebauer, T., Arndt, S.K., Arriga, N., Aubinet, M., Aurela, M., Baldocchi, D., Barr, A., Beamesderfer, E., Marchesini, L.B., Bergeron, O., Beringer, J., Bernhofer, C., Berveiller, D., Billesbach, D., Black, T.A., Blanken, P.D., Bohrer, G., Boike, J., Bolstad, P.V., Bonal, D., Bonnefond, J.M., Bowling, D.R., Bracho, R., Brodeur, J., Brümmer, C., Buchmann, N., Burban, B., Burns, S.P., Buysse, P., Cale, P., Cavagna, M., Cellier, P., Chen, S., Chini, I., Christensen, T.R., Cleverly, J., Collalti, A., Consalvo, C., Cook, B.D., Cook, D., Coursolle, C., Cremonese, E., Curtis, P.S., D'Andrea, E., da Rocha, H., Dai, X., Davis, K.J., De Cinti, B., de Grandcourt, A., De Ligne, A., De Oliveira, R.C., Delpierre, N., Desai, A.R., Di Bella, C.M., di Tommasi, P., Dolman, H., Domingo, F., Dong, G., Dore, S., Duce, P., Dufrêne, E., Dunn, A., Duček, J., Eamus, D., Eichelmann, U., ElKhidir, H.A.M., Eugster, W., Ewenz, C.M., Ewers, B., Famulari, D., Fares, S., Feigenwinter, I., Feitz, A., Fensholt, R., Filippa, G., Fischer, M., Frank, J., Galvagno, M., Gharun, M., Gianelle, D., Gielen, B., Gioli, B., Gitelson, A., Goded, I., Goeckede, M., Goldstein, A.H., Gough, C.M., Goulden, M.L., Graf, A., Griebel, A., Gruening, C., Grünwald, T., Hammerle, A., Han, S., Han, X., Hansen, B.U., Hanson, C., Hatakka, J., He, Y., Hehn, M., Heinesch, B., Hinko-Najera, N., Hörtnagl, L., Hutley, L., Ibrom, A., Ikawa, H., Jackowicz-Korczynski, M., Janouš, D., Jans, W., Jassal, R., Jiang, S., Kato, T., Khomik, M., Klatt, J., Knohl, A., Knox, S., Kobayashi, H., Koerber, G., Kolle, O., Kosugi, Y., Kotani, A., Kowalski, A., Kruijt, B., Kurbatova, J., Kutsch, W.L., Kwon, H., Launiainen, S., Laurila, T., Law, B., Leuning, R., Li, Y., Liddell, M., Limousin, J.M., Lion, M., Liska, A.J., Lohila, A., López-Ballesteros, A., López-Blanco, E., Loubet, B., Loustau, D., Lucas-Moffat, A., Lüers, J., Ma, S., Macfarlane, C., Magliulo, V., Maier, R., Mammarella, I., Manca, G., Marcolla, B., Margolis, H.A., Marras, S., Massman, W., Mastepanov, M., Matamala, R., Matthes, J.H., Mazzenga, F., McCaughey, H., McHugh, I., McMillan, A.M.S., Merbold, L., Meyer, W., Meyers, T., Miller, S.D., Minerbi, S., Moderow, U., Monson, R.K., Montagnani, L., Moore, C.E., Moors, E., Moreaux, V., Moureaux, C., Munger, J.W., Nakai, T., Neirynck, J., Nesic, Z., Nicolini, G., Noormets, A., Northwood, M., Nosetto, M., Nouvellon, Y., Novick, K., Oechel, W., Olesen, J.E., Ourcival, J.M., Papuga, S.A., Parmentier, F.J., Paul-Limoges, E., Pavelka, M., Peichl, M., Pendall, E., Phillips, R.P., Pilegaard, K., Pirk, N., Posse, G., Powell, T., Prasse, H., Prober, S.M., Rambal, S., Rannik, , Raz-Yaseef, N., Reed, D., de Dios, V.R., Restrepo-Coupe, N., Reverter, B.R., Roland, M., Sabbatini, S., Sachs, T., Saleska, S.R., Sánchez-Cañete, E.P., Sanchez-Mejia, Z.M., Schmid, H.P., Schmidt, M., Schneider, K., Schrader, F., Schroder, I., Scott, R.L., Sedlák, P., Serrano-Ortíz, P., Shao, C., Shi, P., Shironya, I., Siebicke, L., Šigut, L., Silberstein, R., Sirca, C., Spano, D., Steinbrecher, R., Stevens, R.M., Sturtevant, C., Suyker, A., Tagesson, T., Takanashi, S., Tang, Y., Tapper, N., Thom, J., Tiedemann, F., Tomassucci, M., Tuovinen, J.P., Urbanski, S., Valentini, R., van der Molen, M., van Gorsel, E., van Huissteden, K., Varlagin, A., Verfaillie, J., Vesala, T., Vincke, C., Vitale, D., Vygodskaya, N., Walker, J.P., Walter-Shea, E., Wang, H., Weber, R., Westermann, S., Wille, C., Wofsy, S., Wohlfahrt, G., Wolf, S., Woodgate, W., Li, Y., Zampedri, R., Zhang, J., Zhou, G., Zona, D., Agarwal, D., Biraud, S., Torn, M., Papale, D., 2020. The FLUXNET2015 dataset and the ONEFlux processing pipeline for eddy covariance data. Scientific Data 7, 225. URL: https://doi.org/10.1038/s41597-020-0534-3, doi:10.1038/s41597-020-0534-3.

Seifi, A., Ehteram, M., Nayebloei, F., Soroush, F., Gharabaghi, B., Torabi Haghighi, A., 2021. Glue uncertainty analysis of hybrid models for predicting hourly soil temperature and application wavelet coherence analysis for correlation with meteorological variables. Soft Computing 25, 10723–10748.

Singhal, M., Gairola, A.C., Singh, N., 2021. Artificial neural network-assisted glacier forefield soil temperature retrieval from temperature measurements. Theoretical and Applied Climatology 143, 1157–1166.

Tuysuzoglu, G., Birant, D., Kiranoglu, V., 2022. Multi-view multi-depth soil temperature prediction (mv-md-stp): a new approach using machine learning and time series methods. International Journal of Intelligent Engineering Informatics 10, 74–104.

Tüysüzoğlu, G., Birant, D., Kiranoglu, V., 2022. Soil temperature prediction via self-training: Izmir case. Journal of Agricultural Sciences 28, 47–62.

Vaswani, A., Shazeer, N., Parmar, N., Uszkoreit, J., Jones, L., Gomez, A.N., Kaiser, Ł., Polosukhin, I., 2017. Attention is all you need. Advances in neural information processing systems 30.

Wang, X., Li, W., Li, Q., 2021. A new embedded estimation model for soil temperature prediction. Scientific Programming 2021, 1–16.

Wang, Y., Zhuang, D., Xu, J., Wang, Y., 2023. Soil temperature prediction based on 1d-cnn-mlp neural network model. Journal of the ASABE , 0.

Woo, G., Liu, C., Sahoo, D., Kumar, A., Hoi, S., 2022. Etsformer: Exponential smoothing transformers for time-series forecasting. arXiv preprint arXiv:2202.01381 .

Zhang, C., Li, Y., Williams, R.A., Chen, Y., Peng, R., Liu, X., Qi, Y., Wang, Z., 2023. Responses of soil respiration and its sensitivities to temperature and precipitation: A meta-analysis. Ecological Informatics , 102057.

Zhou, H., Zhang, S., Peng, J., Zhang, S., Li, J., Xiong, H., Zhang, W., 2021. Informer: Beyond efficient transformer for long sequence time-series forecasting, in: Proceedings of the AAAI conference on artificial intelligence, pp. 11106–11115.